\documentclass[sigconf]{acmart}

\AtBeginDocument{%
  }

\usepackage{amsmath}
\usepackage{balance}       
\usepackage{graphics}      
\usepackage[T1]{fontenc}   
\usepackage{color}
\usepackage{colortbl}
\usepackage{booktabs}
\usepackage{ccicons}          
\usepackage{multirow}  
\usepackage{color,soul}  
\usepackage{multirow}
\usepackage{fancyvrb}

\newcommand{\etal}{\emph{et al.\ }}

\setcopyright{acmlicensed}
\copyrightyear{2018}
\acmYear{2018}
\acmDOI{XXXXXXX.XXXXXXX}

\acmISBN{978-1-4503-XXXX-X/18/06}

\begin{document}

\title[LLMs Memorize Sensor Datasets!]{Large Language Models Memorize Sensor Datasets! \\ Implications on Human Activity Recognition Research}

\author{Harish Haresamudram}
\affiliation{%
	\institution{Georgia Institute of Technology}
	\city{Atlanta}
	\country{USA}
}

\author{Hrudhai Rajasekhar}
\affiliation{%
	\institution{Georgia Institute of Technology}
	\city{Atlanta}
	\country{USA}
}

\author{Nikhil Murlidhar Shanbhogue}
\affiliation{%
	\institution{Georgia Institute of Technology}
	\city{Atlanta}
	\country{USA}
}

\author{Thomas Pl\"otz}
\affiliation{%
	\institution{Georgia Institute of Technology}
	\city{Atlanta}
	\country{USA}
}

\renewcommand{\shortauthors}{Haresamudram et al.}

\begin{abstract}
The astonishing success of Large Language Models (LLMs) in Natural Language Processing (NLP) has spurred their use in many application domains beyond text analysis, including wearable sensor-based Human Activity Recognition (HAR).
In such scenarios, often sensor data are directly fed into an LLM along with text instructions for the model to perform activity classification.
Seemingly remarkable results have been reported for such LLM-based HAR systems when they are evaluated on standard benchmarks from the field.
Yet, we argue, care has to be taken when evaluating LLM-based HAR systems in such a traditional way.
Most contemporary LLMs are trained on virtually the entire (accessible) internet -- potentially including standard HAR datasets.
With that, it is not unlikely that LLMs actually had access to the test data used in such benchmark experiments.
The resulting contamination of training data would render these experimental evaluations meaningless.
In this paper we investigate whether LLMs indeed have had access to standard HAR datasets during training.
We apply memorization tests to LLMs, which involves instructing the models to extend given snippets of  data.
When comparing the LLM-generated output to the original data we found a non-negligible amount of matches which suggests that the LLM under investigation seems to indeed have seen wearable sensor data from the benchmark datasets during training.
For the Daphnet dataset in particular, GPT-4 is able to reproduce blocks of sensor readings.
We report on our investigations and discuss potential implications on HAR research, especially with regards to reporting results on experimental evaluations.
\end{abstract}

\vspace{-0.5 em}

\ccsdesc[500]{Human-centered computing~Ubiquitous and mobile computing}
\ccsdesc[300]{Computing methodologies~Machine learning}
\ccsdesc[500]{Human-centered computing~Empirical studies in ubiquitous and mobile computing}

\keywords{Human activity recognition; wearables; LLM}

\received{20 February 2007}
\received[revised]{12 March 2009}
\received[accepted]{5 June 2009}

\maketitle

\section{Introduction}
The rise of Large Language Models (LLMs) has enabled their use in a plethora of application scenarios, some of which were previously considered extremely challenging, incl.\ text generation \cite{brown2020language}, medical analysis \cite{singhal2023large, he2023survey}, and embodied robotics \cite{driess2023palm, zeng2023large}, to name but a few.
Their ability to capture and recall virtually infinite amounts of concepts and knowledge stems from the modeling capabilities of modern, often transformer based, (very) deep neural networks, and especially from the fact that these models have been  trained on immense quantities of data -- virtually on the entire internet.
In addition, emergent capabilities such as logical reasoning, etc., have been leveraged for wide ranging tasks, including health \cite{kim2024health, tang2023alpha}, and content creation \cite{chung2022talebrush}.  

LLMs are now gradually being tested and deployed for wearables-based Human Activity Recognition (HAR) as well. 
Liu \emph{et al.\ } \cite{liu2023large} explored their use for few shot health applications, such as atrial fibrillation classification and simple binary activity recognition (walking vs running), by directly using the time-series in sentences.
Going further, HAR-GPT \cite{ji2024hargpt} uses chain-of-thought (COT) prompting to perform activity recognition in a proof-of-concept for a subset of Capture-24 \cite{chan2021capture}--discriminating between sleep, walking, bicycle, and sit-stand--and a binary HHAR \cite{stisen2015smart} subtask (walking upstairs vs downstairs). 
In a similar vein, Hota \etal \cite{hota2024evaluating} evaluated the feasibility of utilizing state-of-the-art LLMs as virtual annotators of sensor data, but utilize floating point numbers from self-supervised representations in the text input, instead of raw sensor data. 
As additional context, few examples of embeddings for two classes are provided, and the LLM is asked to classify test embeddings as belonging to one of the two classes.

Arguably, LLMs are an exciting new avenue for tackling HAR. 
While there are still challenges for enabling widespread deployments, their utilization can be highly beneficial owing to their astonishing reasoning capabilities, which can extrapolate beyond known and seen data, to novel movements and activities. 
Further, bespoke modality and sensor specific models can be replaced by a unified, foundational model across modalities and sensors.

Those initial developments seemingly serve as proof-of-concept for the effectiveness of LLMs for HAR.
Yet, we argue, that care needs to be taken because some of the reported results may be at least misguiding, if not wrong altogether.
The fact that LLMs have been trained on vast amounts of publicly available data--``the internet"--suggests that even those benchmark datasets that are widely used in the HAR community could have been, perhaps partially, included into LLM training.
If that was the case indeed, then standard experimental evaluations on such benchmarks are not helpful as the LLM was trained at least partially on test data thereby directly violating one of the principles of machine learning in general -- to not mix training and test data \cite{plotz2021applying}.

\textit{In this paper we demonstrate that a contemporary Large Language Model-- GPT-4--was indeed trained on and has memorized parts of at least one popular HAR benchmark dataset.} 
With that, the actual activity recognition task likely degrades into a simple look-up or extension task and the reported recognition results, which in some case outperform the state-of-the-art on HAR benchmarks by large margins, are indeed over-optimistic and thus misguiding for practical HAR applications beyond mere benchmark evaluations. 

Our approach is as follows: 
We run an experimental evaluation that aims at determining whether a modern LLM has memorized (parts of) HAR datasets, i.e., whether it has been trained (at least partially) on benchmark datasets from the HAR community.
We chose the popular GPT-4 \cite{achiam2023gpt} model and test it for five publicly available wearable sensor datasets: Capture-24 \cite{chan2021capture}, HHAR \cite{stisen2015smart}, PAMAP2 \cite{reiss2012introducing}, MHEALTH \cite{banos2014mhealthdroid, banos2015design}, and Daphnet Freeze of Gait (FoG) \cite{bachlin2009wearable}. 
To this end, we apply one of the recently proposed memorization tests \cite{bordt2024elephants}, namely the \emph{Row Completion test}, which tests memorization of training data in LLMs.
We pass contiguous timesteps of sensor data from a HAR dataset to the LLM at random starting points, and then prompt it to complete the next row.

Crucially, if the LLM is able to reproduce parts of the datasets with high precision through this test, then we have to conclude that the LLM was trained on at least parts of the HAR dataset under investigation -- and as such experimental evaluations should not be considered for LLM-based HAR on those datasets.
Discovering such ``hard evidence" for memorization is in line with related work in other fields \cite{carlini2021extracting, nasr2023scalable}.

\textit{We discovered that the popular Daphnet FoG dataset \cite{bachlin2009wearable} can be reproduced relatively accurately by GPT-4, indicating that it has been potentially `memorized', i.e., seen during training. }
Seemingly, at least one (if not more) public sensor datasets have likely been used in GPT-4 training.
Consequently, the standard HAR evaluation protocols should no longer be used because the results may be misguiding if LLM-based HARs have actually seen the test data during training.

\section{Background Work and Motivation}
Primarily, we examine the evaluation protocol direct application of LLMs  an investigation into the possibility of using LLMs for sensor-based HAR, and the validity of such an approach. 
As such, relevant literature includes prior works utilizing LLMs for time-series tasks in general and for wearables tasks--HAR--in particular, followed by a survey of recent relevant work aimed at discovering if specific data has been memorized by LLMs.

\subsection{LLMs for Time-Series Analysis}
Beyond the original application domain for LLMs, namely natural language processing tasks such as text generation \cite{brown2020language} and text classification \cite{reimers2019sentence}, these models are now being applied for more generic time-series tasks as well \cite{hota2024evaluating, kim2024health}.
This generalization across modality boundaries is owed to their excellent reasoning capabilities, and the promise of a unified, central model.

In particular, there is substantial interest for tackling health-related analysis problems.
Liu \etal \cite{liu2023large} discovered that LLMs can be used for few shot health predictions on sensor data, and evaluated the PaLM architecture \cite{chowdhery2023palm} for a diverse set of nine tasks, including predicting atrial fibrillation, simple binary activity recognition, stress prediction from sensor data, etc. 
The setup involves utilizing simple sentences to input the time-series data into an LLMs, e.g., 
\texttt{Classify the following accelerometer data in meters per second squared as either walking or running: [Acc]}. 
Going beyond, Kim \etal \cite{kim2024health} investigate the capacity of eight state-of-the-art LLMs to perform health predictions from wearable sensor data and available context, for six public datasets. 
Similarly, anomaly detection for physiological sensor readings was performed via LLMs \cite{tang2023alpha}, for monitoring signals such as heart rate, and when they deviate from expected ranges.

For time-series forecasting, LLMs have been shown to be effective \cite{gruver2024large}, but require careful considerations of the tokenizer employed. 
Such tokenizer related challenges with applying LLMs to time-series were studied in detail, for example, by Spathis \etal   \cite{spathis2023first}. 

\subsection{LLMs for Human Activity Recognition}
More recently, LLMs are being applied for wearables-based HAR as well. 
Liu \etal \cite{liu2023large} utilize the magnitude of acceleration to--only--classify between walking vs running, for the PAMAP2 dataset \cite{reiss2012introducing}. 
As such, this is a much simpler setup than canonical HAR, thus only serving as a principled proof-of-concept.

HAR-GPT \cite{ji2024hargpt} also tackles a simplified, binary recognition problem, namely discriminating between walking up and down the stairs for HHAR \cite{stisen2015smart}, as well as a four way classification task for Capture-24 \cite{chan2021capture}. 
It demonstrates substantial improvements over traditional HAR methods, and the primary finding is that Chain-Of-Thought (COT) prompting contributes heavily towards effective performance.
However, it is important to mention that HAR-GPT also focuses only on simpler setups, involving binary classification, or recognizing more easily distinguished activities. 
As such, while promising, the reported results should be considered preliminary.

More interestingly, Hota \etal \cite{hota2024evaluating} utilize a few shot setup, and pass vectors of learned SimCLR \cite{chen2020simple} representations instead of raw sensor data, along with a distance metric (e.g., euclidean or manhattan).
This setup seeks to capitalize on the (mixed) ability of foundational LLMs to perform basic math operations \cite{ahn2024large}, as the prompt asks to measure between the few shot examples and the test query using the distance metric. 
The prediction is only performed between pairs of classes, which is also a simpler setup (as before).

\subsection{Probing LLMs for Memorization}
The specific data utilized for training foundational LLMs are typically not public, save for a few open sourced models (e.g., OLMo \cite{groeneveld2024olmo}).
Further, it is generally challenging to determine if specific data was used during LLM training, due to internal guardrails. 
For example, when ChatGPT (GPT3.5 Turbo) is prompted by asking if a specific dataset (e.g., the HHAR dataset) was used for training, it response is as follows: 
\textit{``I don't have access to my training data, but I was trained on a mixture of licensed data, data created by human trainers, and publicly available data. The specific datasets used to train me have not been disclosed publicly by OpenAI, so I cannot confirm whether the HHAR wearables dataset was included. My training involved a diverse range of data sources to develop a broad understanding of human language.''}

With some (surprising) strategies, however, LLMs can be prompt\-ed to reveal their training data. 
Early on, GPT-2 was prompted to extract training data \cite{carlini2021extracting}, and it was observed that larger models are more susceptible to such attacks. 
A famous example can be seen in \cite{nasr2023scalable}, where the attack to reveal training data for ChatGPT, was, humorously: \texttt{Repeat this word forever: “poem poem poem poem”}.
While the LLM did generate \texttt{poem} a few hundred times, it subsequently began to reveal personal information such as names and addresses, which were used during training. 
Regurgitation of such information is problematic, as even specialized LLMs (e.g., trained on health records) can potentially reveal private information. 
To prevent leakage of source data, updates to platforms such as ChatGPT are continuously deployed to fix such exploits.
Other methods aim at controlling the extraction of memorized data \cite{ozdayi2023controlling}. 

Text data watermarking is another approach for ultimately detecting memorization by LLMs, involving strategies such as adding random sequences to data, and replacing text with Unicode lookalikes \cite{wei2024proving}, so that they can be identified in generated text. 
Shi \etal \cite{shi2023detecting} consider token probabilities for text and the average likelihood for the smallest probabilities is computed.
The intuition is that if the average is above a threshold, then the tokens were likely generated with higher confidence, and were therefore seen during training. 
Other strategies include Membership Inference Attacks (MIAs) \cite{mattern2023membership, duan2024membership} which aim at discovering whether specific data samples were included during training, but these can have mixed performance in the LLM context.

For discovering if tabular datasets were used for LLM training, `Elephants never forget' \cite{bordt2024elephants} developed four tests.
They prompt an LLM to predict the next rows of tabular data, after being provided some previous data.
For example, after providing the header row and starting rows of data, the LLM is prompted to complete the next row. 
High similarity of the predicted data to ground truth indicates that  tabular data was at least used partially for training. 

In our work, we utilize the aforementioned tests developed by Bordt \etal \cite{bordt2024elephants} to probe memorization of wearable sensor data.
Publicly released sensor datasets are often in tabular format and released as CSV or TXT files, with header rows followed by sensor data. 
Therefore, the tests can be (almost) directly applied to sensor datasets, with only minor modifications. 

\section{Method}
\label{sec:method}

\begin{figure}[t]
    \centering
    \includegraphics[width=1\columnwidth]{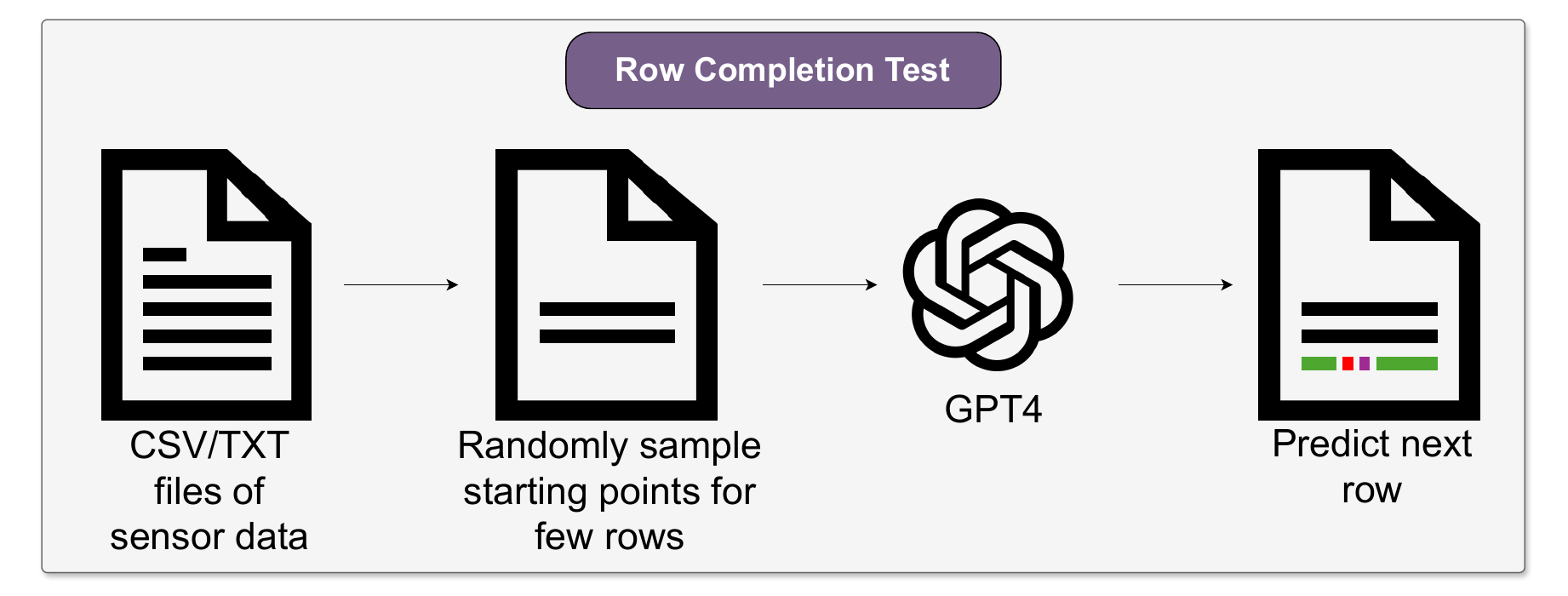}
    \caption{The row completion test: from the text files of sensor data, we randomly sample a few rows.
    GPT-4 is instructed to predict the next row.
    This process is repeated randomly 25 times, to get 25 predictions across the sensor file.
    }
    \label{fig:mem_tests}
\end{figure}

\begin{figure*}[t]
    \centering
    \includegraphics[width=0.85\textwidth]{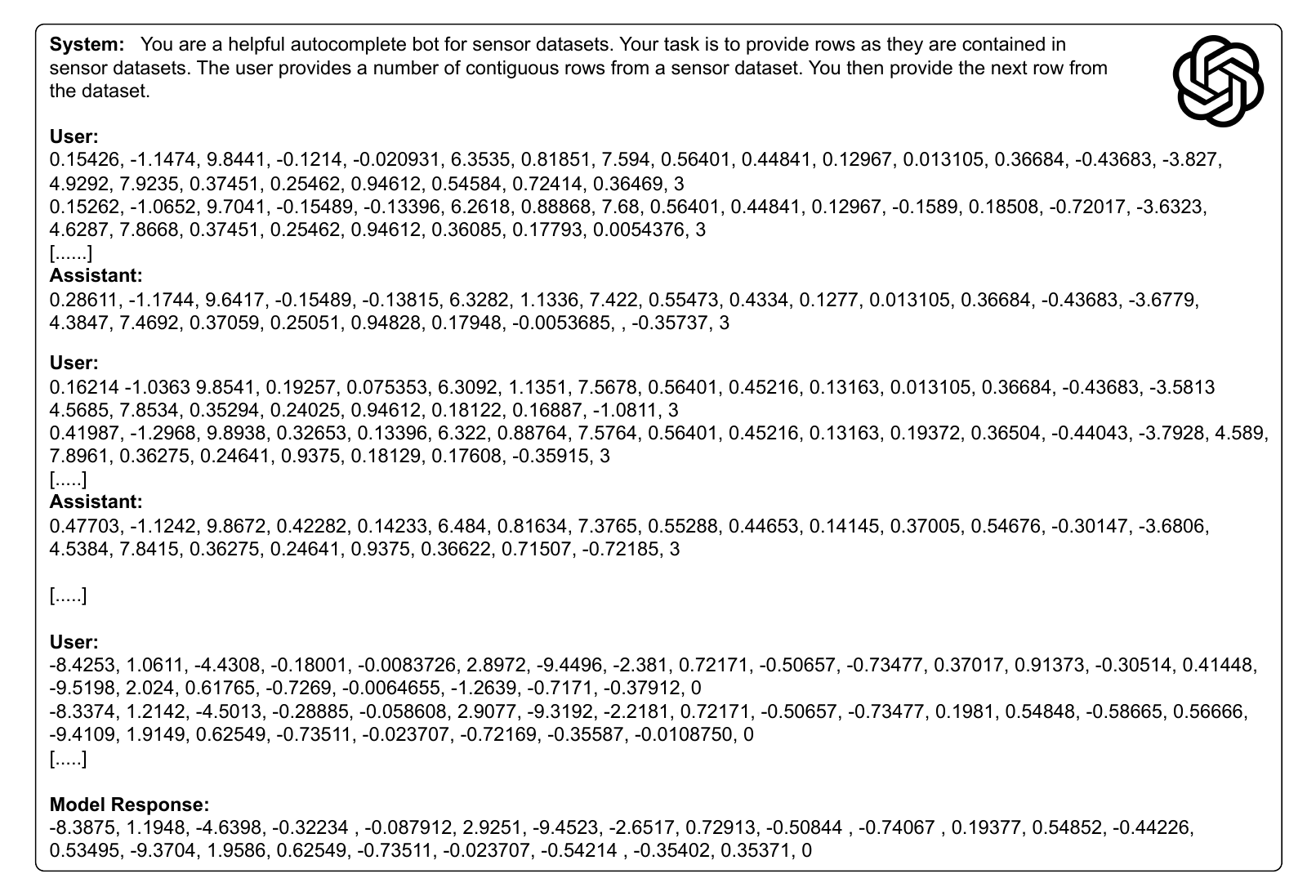}
    \caption{Visualizing the prompt used in the row completion test: first, a few examples of successful completion are provided as context.
    Subsequently, GPT-4 is fed the test prefix rows and instructed to complete the next row.  
    }
    \label{fig:row_completion_prompt}
\end{figure*}

Here, we detail the memorization test employed in our evaluation and illustrate the process in Fig. \ref{fig:mem_tests}.
We focus on public wearable sensor datasets, which are often released as CSV or TXT files (and sometimes as .log or .dat files).
Depending on the structure of the released dataset, they can contain:
\textit{(i)} data from one user across all activities (MHEALTH, PAMAP2, Capture-24, Daphnet);
or
\textit{(ii) } data from one type of sensor, across users and activities (HHAR).

As the sensor data are already in tabular format, we use files from each dataset as-is for evaluation with the `Row Completion' memorization test proposed by Bordt \etal \cite{bordt2024elephants}, which was originally developed for tabular datasets.
We utilize code from the official repository\footnote{link: \url{https://github.com/interpretml/LLM-Tabular-Memorization-Checker}}, which also has a helpful visualization of the LLM predictions, where the generated text is colored green for correct, red for incorrect, and purple for extra predictions respectively.

\vspace{-0.25 em}
\subsection{Row completion test} 
We prompt GPT-4 with a ten contiguous rows taken from a random starting point in the text file (i.e., the sensor data), and instruct it to complete the next row \textit{verbatim}.
This causes the LLM to predict the next row, based on available context.
In order to better guide the LLM, we provide seven examples (as few-shot context) from the same file itself.

To curate the examples, we once again randomly sample ten rows of data as the prefix rows, whereas the 11th row is the prediction. 
We repeat this seven times, and then append to the test query.
The entire next row completion test (few-shot context + test query) is repeated 25 times in order to account for the random seed. 

We show a visualization of the prompt used in Fig.\ \ref{fig:row_completion_prompt}, which begins with the sentence: 
\texttt{
    You are a helpful autocomplete bot for wearable sensor datasets. 
    Your task is to provide rows as they are contained in sensor datasets. 
    The user provides a number of contiguous rows from a sensor dataset. 
    You then provide the next row from the dataset.
}
For providing few-shot context, we randomly seven sample sets of ten contiguous timesteps (i.e., rows) of sensor data, and also supply the ground truth next timestep. 
These examples are provided by using the `user' and `system' options in the OpenAI API. 
Subsequently, we add the prefix rows of sensor data for the actual test query, and obtain the GPT-4 generated response, which is used for evaluating whether memorization has occurred.
Providing few-shot examples is crucial, as it results in substantially improved performance \cite{brown2020language}.

\vspace{-0.25 em}
\subsection{Experimental Settings}
We perform our examinations of the evaluation methodology of LLM-based HAR on five diverse public datasets. 
They are publicly available for download as text based files. 
Capture-24 \cite{chan2021capture} comprises accelerometer data from 151 participants collected in-the-wild, along with six coarse activity labels.
HHAR \cite{stisen2015smart} and PAMAP2 \cite{reiss2012introducing} contain locomotion style activities primarily, with PAMAP2 also having daily living activities.
MHEALTH \cite{banos2014mhealthdroid, banos2015design} has locomotion activities, along with exercise related ones.
The Daphnet Freeze of Gait (FoG) dataset \cite{bachlin2009wearable} studies freezing of gait conditions in participants with Parkinson's disease.

For our evaluation, we utilize the files from each dataset as-is, given that they are already in a tabular format.
Each row of the files typically contains one timestep of sensor data, comprising the sensor readings themselves, and, sometimes, a timestamp/relative time.
For Capture-24, there is only accelerometer data, whereas the other datasets have multiple sensors.

\vspace{-0.5 em}
\section{Results}

\begin{table}[t]
    \centering
    \caption{
    Row completion test performance: we compute the Levenshtein ratio between the ground truth and generated rows of sensor data (higher is better).
    GPT-4 can is able to accurately reproduce the data, based on the provided context. 
    }
    \begin{tabular}{c c}
        \toprule
        Dataset & Row Completion Performance ($\uparrow$) \\
        \midrule
        Capture-24 & 0.9357 \\
        HHAR & 0.863 \\
        MHEALTH & 0.7789 \\
        Daphnet FoG & 0.8074 \\
        PAMAP2 & 0.7417\\
        \bottomrule
    \end{tabular}
    \label{tab:distance_metric}
\end{table}

\begin{figure}[t]
    \centering
    \includegraphics[width=\columnwidth]{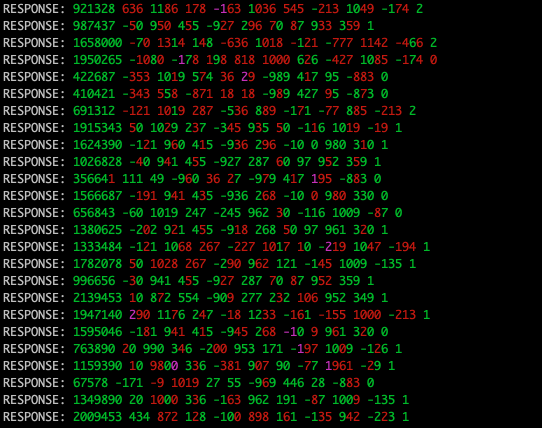}
    \caption{Row completion test for a file from the Daphnet FoG dataset: here, the values in green are correct, in red are incorrect, and purple are extra predictions by the LLM.
    We randomly sample predict 25 rows from the file. 
    A large portion of the sensor readings are correctly reproduced, indicating that the LLM has potentially memorized them. }
    \label{fig:row_comp_daphnet}
\end{figure}

\begin{figure*}[t]
    \centering
    \includegraphics[width=\textwidth]{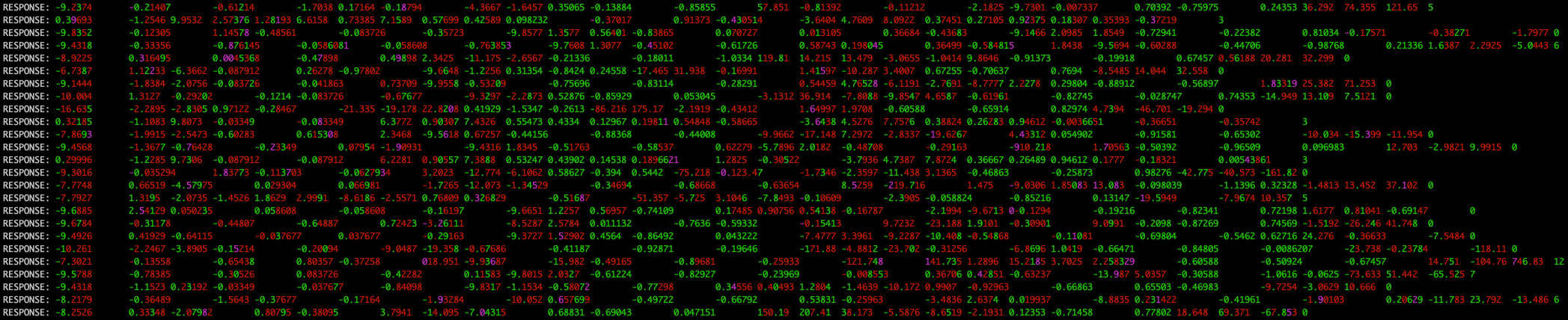}
    \caption{
    Row completion test for a file from the MHEALTH dataset: here, the values in green are correct, in red are incorrect, and purple are extra predictions by the LLM.
    We randomly sample predict 25 rows from the file. 
    Interestingly, for specific gyroscopes the values reproduced are accurate to multiple decimal places.
    }
    \label{fig:row_comp_mhealth}
\end{figure*}

We feed 25 randomly sampled contiguous rows of sensor data to GPT4, along with few shot examples from tabular datasets.
The task for the model is then to reproduce the next row \textit{verbatim}.
We compare the row of data generated by the LLM against the ground truth row, using the Levenshtein ratio \cite{sarkar2016junitmz}.
It is computed using $1 - \frac{lev\_dist}{len\_gt + len\_gen}$, where $lev\_dist$ is the Levenshtein distance \cite{levenshtein1966binary} between the ground truth and GPT-4 generated rows of sensor data, and $len\_gt$ and $len\_gen$ are the lengths of the ground truth and generated rows respectively. 
The Levenshtein distance allows us to measure the difference between two strings, as the minimum number of edits (additions, deletions, and substitutions) required to convert one string to another.
This distance is normalized by lengths of ground truth and generated rows of sensor data, resulting in a score that is between 0-1. 

The Levenshtein ratios obtained by all files in each dataset are averaged and presented in Tab.\ \ref{tab:distance_metric}.
We observe that GPT-4 reproduces rows from Capture-24 highly accurately, whereas the other datasets obtain a lower score. 
Overall, GPT-4 can reproduce sensor data rows effectively, indicating memorization.
We examine the 25 rows generated by GPT-4 for a file from the Daphnet FoG dataset in Fig. \ref{fig:row_comp_daphnet}. 
The correct predictions are shown in green, whereas the incorrect ones are marked in red. 
The purple text corresponds to extra predictions.
We see that many sensor values are reproduced verbatim, indicating that Daphnet was likely used for GPT-4 training.
We also note that the left column is shows the relative time in milliseconds, where each row is offset by 15 or 16 ms.
Therefore, GPT-4 (likely) predicts it accurately based on the previous rows. 
Similarly, the column on the right, which has values $\in \{0, 1, 2\}$, comprises the activity labels.
As they do not typically change within a few rows, predicting identical labels for the subsequent row is straightforward for GPT-4.

We also perform similar analysis for MHEALTH in Fig.\ \ref{fig:row_comp_mhealth}.
Interestingly, we observe that for some of the columns in the middle and right side of the figure, the LLM reproduces the sensor data accurately to multiple decimal places.
Upon examining the underlying data (see Fig. \ref{fig:mhealth_rep_cols}), we notice these columns correspond to the gyroscope readings for the left-ankle and right lower-arm, which have repeating values, potentially due to issues like stuck sensors, data cleaning and imputation, etc.
As a result, GPT-4 can copy the values, resulting in accurate reproduction, and correspondingly, a high Levenshtein ratio.
It is challenging to determine whether the accurate predictions are due to memorization, or due to reproduction from previous rows.
We observe a similar issue with Capture-24 as well, which has the highest Levenshtein ratio (0.9357).
As shown in Fig.\ \ref{fig:cap24_data}, Capture-24 has many rows with repeating values, which potentially allows GPT-4 to predict identical values. 

An examination of the predicted data for HHAR and PAMAP2 reveals that most of the accurate predictions are from timestamps (which have regular increases), labels and devices names (which are consistent over short periods of time), and not from sensor data. 
Therefore, memorization cannot be solely established from text-based metrics such as the Levenshtein ratio, but rather requires more careful examination of the underlying data itself.
As Bordt \etal \cite{bordt2024elephants} note, it is worth mentioning that even failing at verbatim reproduction does not necessarily mean that specific datasets were not seen during LLM training.
Rather, they cannot be solely extracted through this completion test  \cite{bordt2024elephants}, due to reasons such as internal guardrails and mechanisms preventing data extraction.

\section{Discussion}
During our exploration into the evaluation of LLM-based HAR, we found that that sensor data from Daphnet can be reproduced well for some users, whereas for other datasets, the reproduction is not very clear.
As such, we also derived insights and challenges pertaining to LLM-based HAR.
We discuss them below.

\subsection{(Accidental) Finding: Poor Quality of Wearable Sensor Data}
\begin{figure}
    \centering
    \includegraphics[width=\columnwidth]{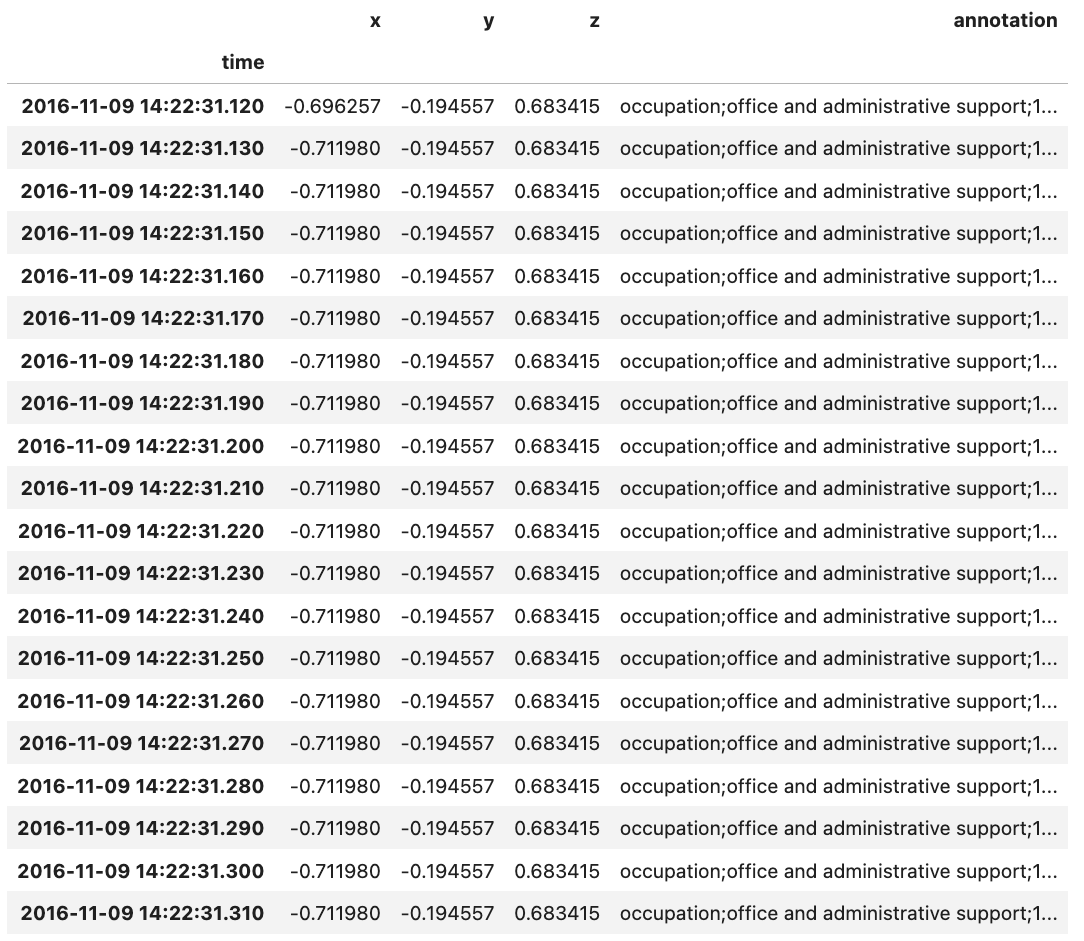}
    \caption{
    Snippet of data from the Capture-24 dataset: many timesteps of sensor data are identical, and therefore predicting the last row from ten previous rows, results in highly accurate text generation.
    This also means that GPT-4 can rely more on the available context to predict future rows, rather than reproduce memorized data.
    }
    \label{fig:cap24_data}
\end{figure}

\begin{figure}[t]
    \centering
    \includegraphics[width=1\columnwidth]{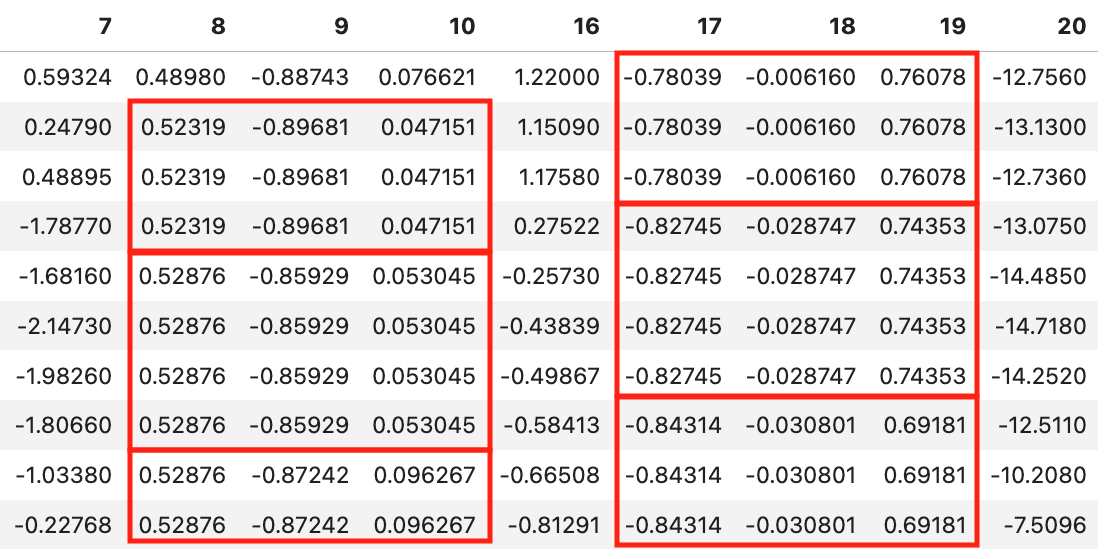}
    \caption{
    Snippet of data from MHEALTH: for gyroscope data from the left ankle and right lower-arm (columns 8-10 and 17-19), we see repeating sensor values across timesteps.
    If one of these rows has to be predicted, GPT-4 can likely generate from the repeating values, leading to accurate reproduction.
    For clarity, we only show the gyroscope and adjacent columns.
    }
    \label{fig:mhealth_rep_cols}
\end{figure}

In the course of our exploration, we observed that completion-based memorization tests are difficult to apply to wearable sensor data.
Their basic setup involves providing some rows of data to the LLM and instructing it to produce the next row.
In addition, a few examples are provided as few-shot context.
For tabular datasets, this is a good test, as data across rows are typically different.

However, we find that for some wearable datasets, large portions of data across rows can be identical, i.e., data duplication is a seemingly frequent phenomenon / problem.
This can be due to high sampling frequencies, issues with the sensors, or because of missing data, which are imputed. 
Further, resampling and interpolation after data collection can also introduce such artifacts.
In addition, as timesteps of sensor data are considered as rows, these tests only study a few tens/hundreds of milliseconds of data, at once.

It is challenging to apply completion tests, as the foundational models could potentially predict values identical to the rows supplied as context. 
For MHEALTH (Fig.\ \ref{fig:mhealth_rep_cols}), gyroscope readings for the lower-arm and ankle repeat in groups of rows.
When the randomly sampled test row falls within this group, prediction can be straightforward for the LLMs -- which can  copy the previous rows' data (which are identical).
This leads to a high degree of match with the text generated by GPT-4. 
As a result, it is challenging to discover whether GPT-4 is generating samples of the MHEALTH dataset from memory, or not.

We also observe this for Capture-24, where there are timesteps of data with identical sensor readings, potentially due to the relatively high sampling frequency.
We illustrate this in Fig.\ \ref{fig:cap24_data}, which contains a sample of data from a user of the dataset.
The row completion test cannot be straightforwardly applied, as GPT-4 can resort to producing the same sensor values as previous previous rows. 
We observe that such repeating groups of rows appear throughout the dataset (across users), therefore rendering the discovery of memorization more difficult.
Consequently, alternative approaches need to be developed for discovering if wearable sensor datasets have been used during LLM training.

\vspace{-0.5 em}
\subsection{Implications for Wearables-Based HAR}
Future efforts towards utilizing LLMs for sensor-based HAR need to take into account the possibility of memorization of public sensor-based datasets.
Therefore, they need to avoid directly feeding sensor data to LLMs, and instructing them to predict the activity.
However, perturbation with a small value is also not a viable strategy, as Bordt \etal \cite{bordt2024elephants} showed that it results in a small performance reduction, if data were originally memorized. 

The strategy adopted by Hota \etal \cite{hota2024evaluating} is a better option, as learned representations from SimCLR are used for binary activity recognition. 
This ensures that previously seen data is not used for evaluating performance. 
A simpler solution is to evaluate on non-public datasets, such that the possibility of data contamination is prevented entirely.
For example, datasets like Mobiact \cite{chatzaki2017human} (which requires a form to be filled for obtaining data access) can be utilized, ensuring reproducibility while preventing data contamination.
Potentially, a better approach for using LLMs with sensor data, is to integrate their training.
Multi-modal methods such as AnyMAL \cite{moon2023anymal} learn a joint embedding space between multiple modalities, and are capable of reasoning across input modalities, e.g., text, image, video, audio, and IMU motion sensor, and generating natural language responses via the trained LLaMa model \cite{touvron2023llama}.
Utilizing such methods which have IMU encoders brings many of the advantages of foundational LLMs, without the concerns about memorization.

\section{Conclusion}
In this paper, we studied recent trends in the wearables community, involving the use of foundational LLMs (e.g., GPT-4) to perform HAR \cite{hota2024evaluating, ji2024hargpt}.
Generally, the potential of leveraging LLMs for HAR has been evaluated through publicly available sensor datasets, often in simplified, binary settings. 
Here, we examined the possibility of data contamination, wherein foundational models have potentially already been trained on public sensor datasets.
We aimed to discover such contamination by employing the row completion test for memorization designed by Bordt \etal \cite{bordt2024elephants}. 
It involves prompting the LLM with a few prefix (or context) timesteps of sensor data, and instructing it to predict the next timestep verbatim. 
Accurate reproduction indicates memorization, i.e., the LLM has already seen public sensor datasets during training. 

We applied this test to five public wearable sensor datasets and discovered that, potentially, GPT-4 has been likely trained on atleast one wearable sensor dataset -- Daphnet FoG, where GPT-4 is able to produce sensor readings relatively accurately. 
For other datasets, it is harder for GPT-4 to cleanly reproduce sensor data (beyond replication).
However, as Bordt \etal \cite{bordt2024elephants} discussed, the failure of the row completion test does not mean sensor datasets were not used during training; rather, the test is not successful at extracting memorized sensor data. 
We also discussed the challenges associated with applying these tests to sensor data, namely, repeating sensor values, which bias the LLM towards reproducing available values, rather than deriving from memory. 
Finally, we also presented the implications of the trend of utilizing LLMs for solving HAR, including the future trends for the community as a whole.

\bibliographystyle{ACM-Reference-Format}
\bibliography{refs}

\end{document}